\documentclass{article}
\usepackage{iftex}
\usepackage{microtype}
\usepackage{graphicx}
\usepackage{booktabs}
\usepackage{enumitem}
\usepackage{xcolor}
\usepackage{colortbl}
\usepackage{hyperref}

\usepackage{amsmath}
\usepackage{amssymb}
\usepackage{mathtools}
\usepackage{amsthm}
\usepackage[capitalize,noabbrev]{cleveref}

\theoremstyle{plain}
\newtheorem{theorem}{Theorem}
\newtheorem{proposition}{Proposition}

\newtheorem{corollary}{Corollary}
\theoremstyle{definition}

\theoremstyle{remark}

\ifPDFTeX\else
  \usepackage{fontspec}
  \AtBeginDocument{\setmainfont{texgyretermes}[Extension=.otf, UprightFont=*-regular, BoldFont=*-bold,
    ItalicFont=*-italic, BoldItalicFont=*-bolditalic]\normalfont}
\fi

\newcommand{\BFChangedRTN}{0.51}
\newcommand{\BFChangedSR}{7.67}
\newcommand{\BFOnce}{57.4}
\newcommand{\BFRTN}{39.9}
\newcommand{\BFTuned}{61.1}
\newcommand{\BitFactorHigh}{5.1}
\newcommand{\BitFactorLow}{3.4}
\newcommand{\BoldLarge}{0.7}
\newcommand{\BoldSmall}{5.7}
\newcommand{\CensusModels}{9}
\newcommand{\ChangedOnce}{0.051}

\newcommand{\ChangedSR}{33.235}
\newcommand{\ClipHalfRate}{28.1}
\newcommand{\ClipNarrow}{29.9}
\newcommand{\ClipWide}{34.4}

\newcommand{\FOOnceBF}{0.38}
\newcommand{\FOOnceThree}{1.56}
\newcommand{\FOSRBF}{10.37}
\newcommand{\FOSRThree}{107.79}
\newcommand{\FreshOnce}{68.0}

\newcommand{\LagPowOnceClip}{0.05}
\newcommand{\LagPowOnceFinal}{68.2}

\newcommand{\LagPowSRClip}{6.33}

\newcommand{\LargeMaxGain}{11.9}

\newcommand{\LargePowGain}{12.3}

\newcommand{\MLXOnce}{54.2}
\newcommand{\MLXSR}{29.6}

\newcommand{\NoiseBitsFive}{0.0018}
\newcommand{\NoiseBitsFour}{0.0085}
\newcommand{\NoiseBitsThree}{0.0306}
\newcommand{\NoiseHigh}{0.037}
\newcommand{\NoiseLarge}{0.0083}
\newcommand{\NoiseLargeOut}{8.5}
\newcommand{\NoiseLow}{0.0081}
\newcommand{\NoiseModels}{9}
\newcommand{\NoiseSmall}{0.0285}
\newcommand{\NoiseSmallOut}{16.3}
\newcommand{\OnceReleaseCount}{33}
\newcommand{\OverlapAvg}{60.7}
\newcommand{\OverlapRatio}{1.26}
\newcommand{\OverlapRatioPlain}{1.29}
\newcommand{\OverlapStepsHr}{154.9}
\newcommand{\PatchOnce}{0.93}

\newcommand{\PatchRatio}{266}
\newcommand{\PatchSR}{248.47}
\newcommand{\SameCellHigh}{99.3}
\newcommand{\SameCellLow}{99.1}
\newcommand{\SmallMaxGain}{11.1}

\newcommand{\SmallPowGain}{19.4}

\newcommand{\StepKLBFOnce}{0.0016}
\newcommand{\StepKLBFSR}{0.0012}
\newcommand{\StepKLFiveOnce}{0.0012}
\newcommand{\StepKLFiveSR}{0.0038}
\newcommand{\StepKLThreeOnce}{0.0009}
\newcommand{\StepKLThreeSR}{0.0573}
\newcommand{\SyncAvg}{53.1}
\newcommand{\SyncStepsHr}{122.7}
\newcommand{\TheoryRunDev}{0.08}
\newcommand{\TheoryRuns}{18}
\newcommand{\TheoryStepDev}{0.70}
\newcommand{\TieMXHigh}{6.20}
\newcommand{\TieMXLow}{6.19}

\newcommand{\TrackOnce}{1.000}
\newcommand{\WireOnce}{2.74}
\newcommand{\WireSR}{1{,}192.20}




\definecolor{oursbg}{HTML}{EEF3FA}
\newcommand{\oursrow}{\rowcolor{oursbg}}

\newcommand{\E}{\mathbb{E}}
\newcommand{\Var}{\operatorname{Var}}
\newcommand{\Prob}{\mathbb{P}}
\newcommand{\rn}{\operatorname{rn}}                 %
\newcommand{\frc}{\operatorname{frac}}
\newcommand{\method}{draw-once rounding}
\newcommand{\Method}{Draw-once rounding}
\newcommand{\ours}{Draw-once (ours)}

\usepackage[preprint]{icml2027}
\hypersetup{pdfsubject={Preprint}}      %
\IfFileExists{release.tex}{\input{release.tex}}{}      %
\newif\ifanonymous
\anonymousfalse

\icmltitlerunning{Draw Once: Coupled Stochastic Rounding for RL Post-Training}

\begin{document}

\twocolumn[
  \icmltitle{Draw Once: Coupled Stochastic Rounding for RL Post-Training}

  \icmlsetsymbol{equal}{*}
  \begin{icmlauthorlist}
  \icmlauthor{Chuyue Wang}{equal,penn}
  \icmlauthor{Shi Wang}{equal,penn}
\end{icmlauthorlist}

\icmlaffiliation{penn}{University of Pennsylvania, Philadelphia, PA, USA}

\icmlcorrespondingauthor{Chuyue Wang}{wangcy07@sas.upenn.edu}

  \icmlkeywords{reinforcement learning, large language models, stochastic rounding, coupling, quantization, weight synchronization, asynchronous training}

  \vskip 0.3in
]

\printAffiliationsAndNotice{\textsuperscript{*}Equal contribution.\newline}

\begin{abstract}
In reinforcement-learning post-training, fresh rounding draws can give a trainer and its sampler different token probabilities at the same checkpoint.
The resulting weight changes consume bandwidth and perturb the likelihood ratios used to train on earlier rollouts.
We propose \emph{\method{}}, which draws one random threshold per weight and reuses it at every export.
For any two fixed master values, this coupling preserves each export's stochastic-rounding distribution and minimizes every convex cost of the exported weight's movement in expectation.
In our experiments with two Qwen3 model sizes and two FP8 grids, the trainer offsets each master weight once at load and then uses its ordinary cast.
With the sampler one export behind, draw-once improves held-out logic-puzzle accuracy over fresh draws by \SmallMaxGain{} to \SmallPowGain{} percentage points on average over training.
On Qwen3-0.6B with power-of-two scales, fresh draws require \PatchRatio{} times as many bytes as draw-once to encode the changed weights.
On four GPUs, overlapping sampling and training with draw-once raises the optimizer step rate to \OverlapRatio{} times its synchronous rate.
\end{abstract}

\section{Introduction}
\label{sec:intro}

Generating answers accounts for much of the cost of reinforcement-learning post-training. Systems split this work between a trainer that updates full-precision master weights and samplers that generate answers from low-precision copies~\citep{schulman2017proximal,shao2024deepseekmath,xi2026jetrl}.
A sampler can use the previous copy while the trainer prepares the next, overlapping two expensive stages~\citep{espeholt2018impala,li2026staleflow,huang2026stable}.
Changed weights must reach the sampler, and differences in token probabilities enter the training objective. Both can arise from rounding alone.

We propose \emph{\method{}}. Each weight draws a random rounding threshold at load and keeps it for every export. The threshold determines which of two neighboring grid values the sampler receives.
Consider exporting an unchanged weight twice. Independent draws can send it to opposite ends of its cell. A kept threshold returns the same value both times, yet gives the usual stochastic-rounding probability of rounding up.
In our logic-puzzle runs with the sampler one export behind, draw-once improves average accuracy over fresh draws on Qwen3-1.7B by \LargePowGain{} points on the power-of-two FP8 grid and \LargeMaxGain{} on the max-abs grid. All three seeds favor the kept threshold on each grid (\cref{tab:main}).

\begin{figure*}[t]
  \centering
  \includegraphics[width=\textwidth]{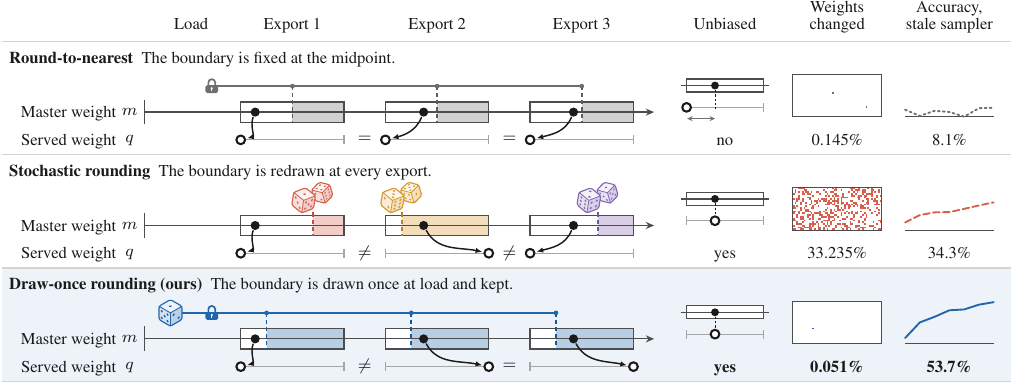}
  \caption{\textbf{A kept threshold changes the served weight only when the master crosses a rounding boundary.} In the illustration, the master (dot) moves less than half a grid cell. The served value (ring) rounds up when the master crosses the dashed boundary. Round-to-nearest fixes that boundary at the midpoint. Stochastic rounding redraws it at each export. Draw-once samples it once per weight. At right, the first column shows the expected served value for a fixed master. The next two show Qwen3-0.6B on the power-of-two FP8 grid with one export of lag. Colored marks give the mean fraction of weights changed, scaled to 1{,}200 weights. Curves show held-out accuracy from 0 to 70\%, with the training average beneath.}
  \label{fig:idea}
\end{figure*}

The served copy must track the master, require little data to synchronize, and provide useful rollouts for the next training step.
Round-to-nearest can make updates sparse. When a released bfloat16 weight already lies on the sampler's grid, a small optimizer step rounds back to the same value. Systems exploit these unchanged weights to cut synchronization volume~\citep{miahi2026understanding,hu2026sparserlsync,peng2026arealdte}, but the sampler can miss many successive updates~\citep{gao2026bf16,nazemi2026closing}.

Stochastic rounding delivers small updates in expectation by drawing between the two neighboring grid values~\citep{gao2026bf16,nazemi2026closing,megatronsr}. Drawing afresh at each export also changes the copy when the master barely moves. On Qwen3-0.6B with power-of-two FP8 scales, a redraw changes about a third of weights per export. Even before training, two redraws of the same checkpoint disagree enough to move sampled tokens outside the trust region of the clipped objective (\cref{fig:static}).
This affects stacks that use FP8 for rollout generation~\citep{xi2026jetrl,qiu2026fp8rl,zhou2026ais} and train on older policy versions~\citep{nazemi2026closing}.

Unbiasedness fixes the distribution of one export. How exports share randomness determines their \emph{coupling}. Suppose a weight moves from a quarter to halfway across a cell. Its draw-once exports differ only if the threshold falls between those positions, which happens with probability one quarter. Draw-once gives every pair the monotone coupling, so expected absolute movement equals the master's displacement (\cref{prop:coupling}). On a uniform grid, one random offset at load implements the rule with the ordinary cast (\cref{fig:idea}).

\newpage %
\noindent\textbf{Contributions.}
\begin{itemize}[leftmargin=1.4em,itemsep=0.25em,parsep=0pt,topsep=0.3em]
  \item We isolate the policy change caused by rounding alone. Across \NoiseModels{} unchanged checkpoints, fresh FP8 exports move sampled tokens outside the clipped objective's ratio band, including \NoiseSmallOut{}\% of tokens on Qwen3-0.6B (\cref{sec:background}).
  \item We propose \emph{\method{}}, which reuses one random threshold per weight across exports. For fixed master values, this coupling preserves each weight's stochastic-rounding law and minimizes every convex cost of its movement at every lag. On a uniform grid, a move of $d\leq1$ cells changes the exported weight with probability $d$ (\cref{prop:coupling,cor:sync}).
  \item We separate learned drift from rounding noise in a local approximation to trainer--sampler policy divergence. Draw-once minimizes the rounding term for fixed master values. With one export of lag, it gains \SmallMaxGain{}--\SmallPowGain{} percentage points over fresh draws in held-out puzzle accuracy averaged over training, with gains at every matched seed (\cref{prop:lag,tab:main}).
  \item We fold each threshold into its master weight at load. The trainer uses its ordinary cast, and samplers reconstruct each export by replacing only weights that cross a rounding boundary. On Qwen3-0.6B with power-of-two scales, this cuts encoded patch size \PatchRatio{}-fold relative to fresh draws. On four GPUs, overlapping sampling and training with draw-once yields \OverlapRatio{} times the synchronous step rate (\cref{alg:drawonce,sec:sync}).
\end{itemize}

\section{Background and Motivation}
\label{sec:background}

A released bfloat16 weight lies exactly on a grid point when the sampler also uses bfloat16. On a coarser grid it can lie between two points, even on a rounding tie. These starting positions explain why the same optimizer update can disappear in one format and move a served weight by a full cell in another.

\subsection{The Export Step}
\label{sec:export}

Consider one weight.
The optimizer holds a float32 master $m_t$, and after step $t$ the samplers receive a served value $q_t$ on a coarser grid.
Near the weight the grid has \emph{cell} width $s$.
The bfloat16 format keeps 7 of float32's 23 mantissa bits, so $s$ is $2^{-7}$ of the weight's binade, and FP8 (E4M3) keeps 3, so its cell is 16 times wider.
An FP8 tensor also carries a scale, either a power of two shared by a small block of weights, as in the MX formats~\citep{rouhani2023microscaling}, or the tensor's largest magnitude.
We write $\phi_t = \frc(m_t / s)$ for the \emph{phase}, the master's position inside its cell, and $d_t = |m_{t+1} - m_t| / s$ for the \emph{move}, one optimizer step measured in cells.
At a learning rate of $10^{-6}$ a typical move on bfloat16 is about a hundredth of a cell.

A served copy is \emph{unbiased} when $\E[q_t] = m_t$.
A sampler has \emph{lag} $L$ when it generates from $q_{t-L}$ while the trainer updates $m_t$.
With one export of lag, the clipped objective compares the current and previous copies' probabilities for the same token given the same prefix.
The \emph{synchronization} of an export ships the weights that changed since the previous one, as a sparse patch.

\begin{figure}[t]
  \centering
  \includegraphics[width=\columnwidth]{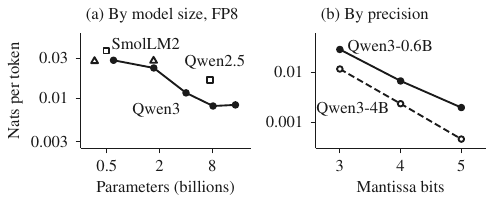}
  \caption{\textbf{Two stochastic-rounding exports of one unchanged checkpoint are two different policies.} Divergence between their policies in nats per sampled token, with the linear layers at the sampler's precision; each point is the mean over six ordered pairs of three exports. (a) At FP8 precision, for \NoiseModels{} models of three families. (b) By the sampler's mantissa bits, for two models; a bit dropped multiplies the divergence \BitFactorLow{} to \BitFactorHigh{} times, where the rounding term of \cref{prop:lag} gives four.}
  \label{fig:static}
\end{figure}

\subsection{Round-to-Nearest Withholds Updates}
\label{sec:rtn-bg}

Round-to-nearest sets $q_t = \rn(m_t)$, the grid value closest to the master.
When the released checkpoint and serving grid are both bfloat16, $m_0$ is itself a grid value.
Every weight starts half a cell from the nearest rounding boundary, and its served value stays at $q_0$ until the accumulated update exceeds half a cell.
\citet{gao2026bf16} measure that only 0.4 to 0.5\% of served weights change per step this way and that the sampler runs many steps behind its master, and \citet{nazemi2026closing} trace the slower convergence of bfloat16 training to the same cause.
Sparse synchronization rests on the same fact from the other side.
\citet{miahi2026understanding} and \citet{hu2026sparserlsync} report that about 99\% of weights are unchanged by a step after the cast, and ship only the rest.

Moving from a bfloat16 checkpoint to a coarser serving grid creates a second effect.
A bfloat16 value carries 7 mantissa bits and E4M3 keeps 3, so when the FP8 scale is a power of two one weight in sixteen has its four dropped bits equal to $1000_2$ and sits exactly on a rounding tie.
\cref{app:static} counts them in released checkpoints, at \TieMXLow\% to \TieMXHigh\% of all matrix weights under MXFP8 in each of \CensusModels{} checkpoints from 0.36B to 14B parameters.
At a tie, an arbitrarily small update toward the other neighbour can flip the served weight by a whole cell.

\subsection{A Redraw Moves the Policy and Fills the Patch}
\label{sec:redraw}

Stochastic rounding draws $u_t \sim U[0,1)$ for each weight at each export and sets $q_t = s\lfloor m_t / s + u_t \rfloor$, the upper neighbour with probability $\phi_t$.
Each export is unbiased.
Two exports of masters with phases $\phi$ and $\phi'$ in one cell disagree with probability
\begin{equation}
  \phi(1 - \phi') + \phi'(1 - \phi) \;=\; 2\phi(1-\phi) + (\phi' - \phi)(1 - 2\phi).
  \label{eq:redraw}
\end{equation}
The first term does not contain the update.
It is the price of drawing again, it is paid at every export, and once the phases are spread over the cell it averages to one third.

The price has two sides, and both can be measured without training.
On the weight side, a third of the weights change at every export on an FP8 grid, where the phases of a bfloat16 checkpoint are already spread.
We measure \ChangedSR\% on average in the 0.6B power-of-two runs.
On bfloat16 the phases start together and spread as training proceeds, so the share grows with the length of the run, to \BFChangedSR\% per step over our 60 steps against \BFChangedRTN\% for round-to-nearest.
\citet{gao2026bf16} report 2 to 15\% in their runs and estimate that a patch of changed weights would compress 5 to 30 times less.
A pull request to Megatron-LM reports 19.8\% with stochastic rounding against 1.3\% with round-to-nearest in a 30B-parameter run~\citep{megatronsr}.
It takes its random bits from a hash of each weight's float32 pattern, so the rounding changes whenever the master does.

On the policy side, we export one unchanged checkpoint several times and compare the policies of the copies on the tokens they sample (\cref{fig:static}).
With the linear layers at FP8 precision, as in a released FP8 model, two exports of Qwen3-0.6B are \NoiseSmall{} nats per token apart and \NoiseSmallOut\% of sampled tokens have a probability ratio outside the band $[0.8, 1.2]$ in which PPO-style objectives trust their data.
The two largest Qwen3 checkpoints have similar policy gaps.
At 14B it is \NoiseLarge{} nats per token with \NoiseLargeOut\% of tokens outside the band.
Across \NoiseModels{} models of three families it ranges from \NoiseLow{} to \NoiseHigh{} (\cref{app:static}).
These measurements isolate the change in token probabilities that a lagging sampler can see before the optimizer takes a step.

\section{Draw-Once Rounding}
\label{sec:method}

A single threshold can preserve the rounding law of each fixed weight and minimize how far its rounded value moves across exports.

\subsection{Exports as Couplings}
\label{sec:couplings}

An \emph{export rule} turns the masters $m_0, m_1, \dots$ of a weight into served values $q_0, q_1, \dots$ with randomness of its own.
Two things about the rule matter to a training system, and they are separate.

The first is its \emph{marginal}, the law of one export $q_t$ given $m_t$.
The rule is unbiased when $\E[q_t] = m_t$, and if it serves one of the two grid values next to the master, the marginal is then determined.
It is the law $\mathrm{SR}(m_t)$ of a stochastic rounding, the upper neighbour with probability $\phi_t$ and the lower one otherwise.

The second is its \emph{coupling}, the joint law of two exports $(q_t, q_{t'})$ with those marginals.
When each export draws its own randomness the coupling is the independent one, and when they share randomness we say that the exports are \emph{coupled}.
A \emph{coupled stochastic rounding} is a rule whose every export has the law $\mathrm{SR}(m_t)$ and whose exports are coupled.
What a system pays for between two exports is a function of the difference $q_{t'} - q_t$.
A synchronization ships its non-zero entries, and a sampler that lags by $t' - t$ exports generates from a policy that differs from the trainer's by it (\cref{sec:lagtheory}).

Unbiasedness constrains the marginal and leaves the coupling free.
\cref{tab:theory} places the rules of \cref{sec:background} on these two axes.
Round-to-nearest from a released checkpoint has a degenerate marginal and a rigid coupling.
A redraw has the unbiased marginal and the independent coupling. For uniformly spread phases, its probability of changing a weight approaches one third as the update tends to zero.
\Method{} is the coupled stochastic rounding whose exports share all of their randomness.
It attains the lower bounds on expected movement and rounding variance.

\begin{table*}[t]
  \caption{\textbf{Unbiasedness fixes one export, and the coupling fixes what moves between two.} Two exports $q$ and $q'$ whose masters are $d \le 1$ cells apart on a grid of cell $s$, with phases spread over the cell (\cref{prop:coupling,cor:sync,prop:lag}). Round-to-nearest is taken from a released checkpoint, where nothing changes until the accumulated move passes half a cell, and the entries of stochastic rounding are their limits for a small move, which do not vanish with it. The last row bounds every export with unbiased snapshots.}
  \label{tab:theory}
  \centering
  \small
  \begin{tabular}{lllccc}
    \toprule
    & One export & \multicolumn{4}{c}{Two exports} \\
    \cmidrule(lr){2-2}\cmidrule(l){3-6}
    Export rule & Unbiased & Coupling & \shortstack{Weight changes\\$\Prob(q' \neq q)$} & \shortstack{Movement\\$\E|q' - q|$} & \shortstack{Rounding variance\\$\Var(q' - q)$} \\
    \midrule
    Round-to-nearest & no & deterministic & $0$ & $0$ & $0$ \\
    Stochastic rounding & yes & independent & $\tfrac13$ & $\tfrac13\,s$ & $\tfrac13\,s^2$ \\
    \oursrow \textbf{\ours} & yes & monotone & $d$ & $d\,s$ & $d(1-d)\,s^2$ \\
    \addlinespace
    Any unbiased export & yes & any & $\ge d - \tfrac14 d^2$ & $\ge d\,s$ & $\ge d(1-d)\,s^2$ \\
    \bottomrule
  \end{tabular}
\end{table*}

\subsection{The Rule}
\label{sec:rule}

Stochastic rounding compares the phase of each weight with a fresh uniform number.
\emph{\Method{}} compares it with the same number every time.
Each weight draws its own threshold $r \sim U[0,1)$ once, when training starts, and keeps it for every export.
In the notation of \cref{sec:export} the three rules serve
\begin{align}
  \text{round-to-nearest:} && q_t &= s \left\lfloor \frac{m_t}{s} + \frac12 \right\rfloor , \label{eq:rtn} \\
  \text{stochastic rounding:} && q_t &= s \left\lfloor \frac{m_t}{s} + u_t \right\rfloor , \label{eq:sr} \\
  \text{\method{}:} && q_t &= s \left\lfloor \frac{m_t}{s} + r \right\rfloor . \label{eq:threshold}
\end{align}
They differ only in the number added before the floor, which is a constant, a draw $u_t \sim U[0,1)$ made anew for each weight at every export, and the kept draw $r$.
Each rule serves the lower neighbour while the master's phase is below a boundary and the upper neighbour otherwise.
The boundary is the midpoint in \cref{eq:rtn}, a new point $1 - u_t$ at every export in \cref{eq:sr}, and the kept point $1 - r$ in \cref{eq:threshold}, which are the three lanes of \cref{fig:idea}.
\cref{eq:sr,eq:threshold} both add a uniform number, which gives a single export the same law under either rule (\cref{prop:unbiased}).
They differ between two exports, where $u_t$ and $u_{t'}$ are independent and $r$ is shared.
Under \cref{eq:threshold} the served value therefore moves up one cell when the master's phase passes $1 - r$, and moves back when it returns.

On a uniform serving grid the threshold can be folded into the master.
Add the offset $o = (r - \tfrac12)\,s$ to the master once and round to nearest from then on:
\begin{equation}
  m_0 \leftarrow m_0 + o, \qquad q_t = \rn(m_t) = s \left\lfloor \frac{m_t}{s} + \frac12 \right\rfloor .
  \label{eq:offset}
\end{equation}
\cref{eq:offset} is \cref{eq:threshold} with the threshold folded into the master, and \cref{alg:drawonce} states it.
With a sampler in the released format the first export is the released checkpoint itself, because the offset is smaller than half a cell.
The optimizer carries the offset forward with the weights, and four properties come with it.
The export is the cast the format already defines, which is \cref{eq:rtn} up to its rule for an exact tie, so the patch format and the synchronization path are unchanged.
No per-weight state is kept beside the master, and no random number is drawn after load.
The served copy is a function of the master alone, so ranks, replicas and a restarted job agree without shared generators.
And the saved float32 checkpoint is within half a cell of the served copy at every weight, so it can be cast to any release format directly (\cref{sec:rtn}).

\begin{algorithm}[t]
  \caption{\Method{} (offset form)}
  \label{alg:drawonce}
  \begin{algorithmic}[1]
    \REQUIRE released checkpoint $m$; cell width $s(\cdot)$ of the released format; lag $L \ge 0$
    \FOR{each weight $i$ \textbf{once, at load}}
      \STATE $m_i \leftarrow m_i + (r_i - \tfrac12)\, s(m_i)$, \ $r_i \sim U[0,1)$
    \ENDFOR
    \FOR{$t = 0, 1, 2, \dots$}
      \STATE $q_t \leftarrow \rn(m)$ \hfill $\triangleright$ the format's own cast
      \STATE send $q_0$ at load, then ship changed weights
      \STATE take rollouts sampled from $q_{\max(0,t-L)}$
      \STATE $g \leftarrow \nabla \mathcal{L}(q_t)$ on those rollouts
      \STATE $m \leftarrow \textsc{OptimizerStep}(m, g)$
    \ENDFOR
  \end{algorithmic}
\end{algorithm}

The fixed-grid equivalence uses an offset as wide as one serving cell. Our FP8 training runs instead draw the offset within the narrower bfloat16 release cell, so the effective phases on the FP8 grid also depend on the checkpoint.
The low mantissa bits of the checkpoint spread weights across each wider FP8 cell, and the offset disperses weights initially concentrated on its ties (\cref{sec:rtn-bg}).
A stack that already ships stochastic rounding obtains \cref{eq:threshold} by freezing its draw.
Where the reference implementation adds 16 random bits to the float32 pattern and truncates~\citep{gao2026bf16}, \method{} keeps those bits fixed per weight, from load or from the end of a warm-up (\cref{app:late}).

\subsection{Each Export Is Unbiased}
\label{sec:unbiased}

Fix a weight and treat the grid near it as uniform.

\begin{proposition}[One export]
\label{prop:unbiased}
Let $m_t$ not depend on the threshold $r$.
Then the export~\eqref{eq:threshold} has the law $\mathrm{SR}(m_t)$ at every $t$, and in particular $\E[q_t] = m_t$.
Conversely, let $o$ be a random offset supported in one cell.
If round-to-nearest of $m + o$ is unbiased for every $m$, then $o$ is uniform on the cell.
\end{proposition}

At any single export \method{} is therefore indistinguishable from stochastic rounding, and the uniform offset is the only starting position for which the format's ordinary cast passes updates of every size without bias.
A released checkpoint is the opposite extreme, with all of its phase mass at one point.
The first half of the proposition is the dither identity of signal processing~\citep{schuchman1964dither,wannamaker2000theory}.
The difference here is that the dither is drawn once and then travels inside the signal.

In training the master does depend on $r$, through gradients taken at the served copy, so we test the change rates predicted for fixed master moves.
The run-average share of changed weights stays within \TheoryRunDev\% of the value that independence predicts in each of \TheoryRuns{} runs (\cref{sec:sync}), and the served copy's displacement is \TrackOnce{} of the master's.
For Qwen3-0.6B, resampling the thresholds at evaluation gives \FreshOnce\% against \LagPowOnceFinal\%.
At seed 0 the predicted relative standard deviation of the first-order change delivered to the sampler is \FOOnceThree\% at FP8 precision, where a redraw of the same masters gives \FOSRThree\%.
The rounding error is averaged across the weights of one export instead of across exports in time (\cref{app:firstorder}).

\subsection{Least Movement Between Exports}
\label{sec:coupling}

Consider two exports, of masters $m$ and $m'$, any number of steps apart.
By \cref{sec:couplings} an unbiased rule is a choice of coupling of $\mathrm{SR}(m)$ and $\mathrm{SR}(m')$, and a redraw chooses the independent one.
The next result identifies the best choice.

\begin{theorem}[Two exports]
\label{prop:coupling}
Let $(q, q')$ be exports of $m$ and $m'$ under any rule whose snapshots have the laws $\mathrm{SR}(m)$ and $\mathrm{SR}(m')$, and let $(q_\star, q'_\star)$ be the exports of \method{}.
Then for every convex $c$,
\begin{equation}
  \E\, c(q'_\star - q_\star) \;\le\; \E\, c(q' - q),
\end{equation}
and for strictly convex $c$ equality holds only if $(q, q')$ has the joint law of $(q_\star, q'_\star)$.
As special cases, $\E\,|q'_\star - q_\star| = |m' - m|$, and $\E\,(q'_\star - q_\star)^2 = s\,|m' - m|$ for a move of at most one cell.
\end{theorem}

As a function of the kept threshold, $q_\star$ is the quantile function of $\mathrm{SR}(m)$, so one threshold for both exports is the monotone coupling of the two laws, which is optimal on the line for every convex cost~\citep{villani2003topics,santambrogio2015optimal}.
Its expected absolute movement equals the displacement of the master.
Under a redraw a weight inside a cell travels $s\,[\phi(1-\phi') + \phi'(1-\phi)]$ in expectation, about a third of a cell, even when the master did not move.

A sampler can skip an export. Consider a master that goes from $1/4$ to $3/4$ and back to $1/4$ on the unit grid $\{0,1\}$. The least possible probability of a change on either step is $1/2$. A rule can attain both one-step minima by flipping two thirds of the zeros on the way up and, with fresh randomness, two thirds of the ones on the way down. Its first and last exports still differ with probability $1/3$.
Draw-once attains the same one-step minima and returns to the original served value for every threshold. One draw thus realizes the optimal coupling for every pair of exports in a fixed master sequence, however many versions the sampler skips.

The number of changed weights is not a convex cost, so it has a statement of its own.

\begin{corollary}[Synchronization]
\label{cor:sync}
Under \method{} a weight changes between two exports with probability $\min(1, d)$, where $d = |m' - m|/s$.
No rule with unbiased two-neighbour snapshots changes it with probability below the total-variation distance of the two laws, which for a uniform starting phase and $0 \le d \le 1$ averages $d - d^2/4$.
\end{corollary}

For a move of a hundredth of a cell the difference is 0.25\% of the draw-once change rate, and its patch shrinks with the learning rate, as the patch of round-to-nearest does.
Under a redraw the changed share tends to a third whatever the learning rate (\cref{eq:redraw}).
The same optimality appears in two other fields that share randomness to keep two random selections close: survey sampling keeps a ``permanent random number'' per unit to hold successive samples together~\citep{brewer1972selecting,matei2005maximal}, and replicable learning rounds under shared randomness~\citep{kalavasis2024replicable}.

\subsection{What a Lagging Sampler Sees}
\label{sec:lagtheory}

Let the sampler hold the export of $m$ and the trainer the export of $m'$, write $\Delta = q' - q$ and $D = m' - m$ over all weights, and fix the Fisher matrix $F$ at a reference policy independent of the rounding draws.
The local quadratic approximation to the policy divergence uses $\tfrac12 \Delta^{\!\top} F \Delta$.

\begin{proposition}[Drift and rounding]
\label{prop:lag}
For a rule that rounds weights independently with unbiased snapshots,
\begin{equation}
  \E\!\left[\Delta^{\!\top} F \Delta\right] \;=\; \underbrace{D^{\top} F D}_{\text{learned drift}} \;+\; \underbrace{\textstyle\sum_i F_{ii} \Var(\Delta_i)}_{\text{rounding}} .
  \label{eq:lag}
\end{equation}
Under \method{} $\Var(\Delta_i) = s_i^2\, d_i (1 - d_i)$ for $d_i \le 1$, the least possible.
Under a redraw $\Var(\Delta_i) = s_i^2\,[\phi_i(1-\phi_i) + \phi'_i(1-\phi'_i)]$, which averages $s_i^2/3$ for spread phases.
\end{proposition}

The drift term is the same for every unbiased export.
It measures the displacement of the two masters in the local policy geometry. A kept threshold minimizes the additional variance from rounding. For small moves this variance shrinks with the move and vanishes at zero. Independent redraws of an unchanged weight still have variance $2s_i^2\phi_i(1-\phi_i)$ in their difference. At uniformly spread phases, doubling the cell width multiplies the average redraw variance by four.

We also measure consecutive policies on a fixed prompt set during the first ten warm-up steps in MLX. At FP8 precision they differ by \StepKLThreeSR{} nats per token under a redraw and \StepKLThreeOnce{} under draw-once. With two more mantissa bits the distances are \StepKLFiveSR{} and \StepKLFiveOnce{}, respectively. On bfloat16 they are \StepKLBFSR{} and \StepKLBFOnce{}.

An objective that forms a ratio between the trainer's policy and the sampler's cannot tell the two terms apart.
For the clipped surrogate of PPO and GRPO the consequence is specific.
Outside the clip band, the gradient has a component whose direction is independent of the advantage sign and points back toward the sampler (\cref{app:clip}).
When the ratio is drift, this is the trust region at work.
When it is the noise of a redraw, the policy is pulled toward one random draw at this step and another at the next, even for a negligible master update (\cref{app:clip}).

\section{Experiments}
\label{sec:experiments}

\begin{table*}[t]
  \caption{\textbf{With a sampler one export behind, a kept threshold beats fresh draws on both FP8 grids at both sizes.} Held-out accuracy on knights-and-knaves in percent, averaged over training, with final accuracy in parentheses. Each cell averages three seeds. In the lagged columns, bold includes means within half the largest draw-once seed range on that model's two grids (\BoldSmall{} points at 0.6B, \BoldLarge{} at 1.7B).}
  \label{tab:main}
  \centering\small
  \begin{tabular}{l rrr rr}
    \toprule
    & \multicolumn{3}{c}{One export behind} & \multicolumn{2}{c}{No lag} \\
    \cmidrule(lr){2-4}\cmidrule(l){5-6}
    FP8 scales & Round-to-nearest & Stochastic & \textbf{Draw-once} & Stochastic & \textbf{Draw-once} \\
    \midrule
    \multicolumn{6}{l}{\emph{Qwen3-0.6B}} \\
Power-of-two & 8.1 (15.0) & 34.3 (45.8) & \textbf{53.7} (68.2) & 54.8 (65.1) & 57.5 (71.2) \\
Max-abs & \textbf{54.1} (68.8) & 40.2 (49.2) & \textbf{51.3} (67.2) & 54.6 (69.1) & 52.0 (68.2) \\
\addlinespace
\multicolumn{6}{l}{\emph{Qwen3-1.7B}} \\
Power-of-two & 7.8 (9.9) & 59.3 (71.5) & \textbf{71.6} (87.5) & 73.6 (90.4) & 77.6 (90.5) \\
Max-abs & \textbf{71.7} (86.8) & 58.9 (70.7) & 70.8 (85.8) & 78.9 (91.8) & 77.1 (88.8) \\
 
    \bottomrule
  \end{tabular}
\end{table*}

\begin{figure*}[t]
  \centering
\begin{minipage}[t]{0.485\textwidth}
  \centering\includegraphics[width=\linewidth]{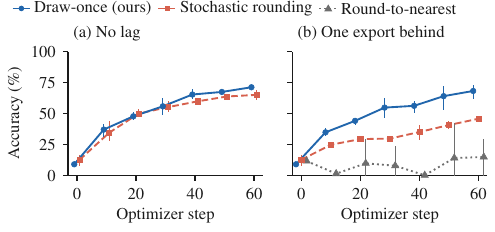}
  \caption{\textbf{On Qwen3-0.6B, draw-once gains \SmallPowGain{} points of average accuracy over fresh draws with one export of lag.} Greedy held-out accuracy with power-of-two scales. Points are means of three seeds and bars span their range.}
  \label{fig:curves}
\end{minipage}\hfill
\begin{minipage}[t]{0.485\textwidth}
  \centering\includegraphics[width=\linewidth]{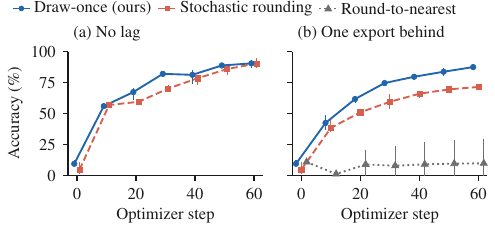}
  \caption{\textbf{On Qwen3-1.7B, draw-once leads a redraw at every evaluation after training starts when the sampler lags one export.} Accuracy of Qwen3-1.7B on the same task and grid as \cref{fig:curves}. Each point averages three seeds and bars span their range.}
  \label{fig:curves-large}
\end{minipage}
\end{figure*}

\subsection{Experimental Setup}
\label{sec:setup}

We train all weights of Qwen3-0.6B and Qwen3-1.7B~\citep{yang2025qwen3} on knights-and-knaves, a logic puzzle with verifiable answers from Reasoning Gym~\citep{stojanovski2025reasoning}.
Following the served-copy training recipe of \citet{gao2026bf16}, we use GRPO~\citep{shao2024deepseekmath} with 24 prompts and eight answers per prompt. AdamW~\citep{loshchilov2019decoupled} uses a learning rate of $10^{-6}$, a 20-step linear warm-up, weight decay 0.01 and gradient clipping at norm 1.
The default objective is the clipped surrogate with $\epsilon=0.2$.
Every ten steps we score greedy answers to 256 held-out puzzles. Average accuracy is the mean of the six scores from steps 10 through 60. Final accuracy is the score at step 60.

Each FP8-precision sampler rounds its matrix weights to three mantissa bits, with either power-of-two scales or a max-abs scale per tensor. Other tensors remain in bfloat16.
The comparisons vary the export rule (\cref{sec:rule}) and whether the sampler uses the current or previous export.
Results are from PyTorch except where we identify an independent MLX measurement. Each main accuracy cell averages seeds 0, 1 and 2. \Cref{app:setup} gives the hardware, formats and timing protocol.

\subsection{Stale Samplers}
\label{sec:stale}

With the sampler one export behind, draw-once improves average accuracy over a redraw by \SmallPowGain{} points on Qwen3-0.6B and \LargePowGain{} points on Qwen3-1.7B on the power-of-two grid (\cref{tab:main}).
On the max-abs grid the gains are \SmallMaxGain{} and \LargeMaxGain{} points, respectively. Across all four comparisons, the lowest draw-once seed exceeds the highest redraw seed (\cref{app:more}). The power-of-two learning curves show when the advantage emerges. At both sizes, draw-once leads fresh draws at every held-out evaluation after the initial export with the sampler one export behind (\cref{fig:curves,fig:curves-large}).
An independent MLX implementation reproduces the lagged comparison on Qwen3-0.6B with power-of-two scales: \MLXOnce\% for draw-once and \MLXSR\% for a redraw, over three seeds.

\begin{figure}[t]
  \centering\includegraphics[width=\columnwidth]{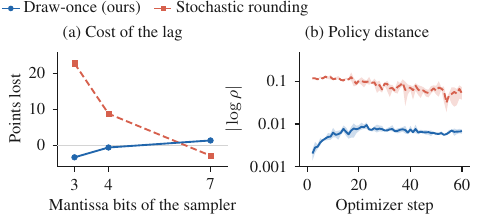}
  \caption{\textbf{Coarser rounding makes stale rollouts costlier under a redraw.} Qwen3-0.6B. (a) Average accuracy with no lag minus that one export behind at matched seed 0. (b) Mean absolute token log-ratio between trainer and sampler on the three-bit grid, one export behind. Lines average three seeds and bands span their range.}
  \label{fig:mech}
\end{figure}

The two rounding rules produce different likelihood ratios in the clipped objective (\cref{fig:mech}). On Qwen3-0.6B with power-of-two scales, clipping zeroes the policy-gradient contribution of \LagPowSRClip\% of response tokens under a redraw and \LagPowOnceClip\% under draw-once, averaging the per-step fractions over three seeds.
The accuracy gap narrows as the serving grid becomes finer. With token-level importance weights or a plain policy gradient, the gap also narrows on the original FP8 grid (\cref{app:depth}).
On integer arithmetic, a kept threshold again outperforms a redraw one export behind in both model families we tested (\cref{app:more}).

\subsection{Synchronization and Overlapped Training}
\label{sec:sync}

\begin{table}[t]
  \caption{\textbf{Draw-once exports stay sparse at both sizes.} One export behind on the power-of-two grid, three seeds. Weight changes average all 60 exports. Patch sizes use a common encoding. Single-GPU export times average the last 55 exports per seed (\cref{app:setup}).}
  \label{tab:sync}
  \centering\small
  \begin{tabular}{lrrr}
    \toprule
    Export rule & \shortstack[r]{Weights\\changed (\%)} & \shortstack[r]{Patch\\(MB)} & Export (s) \\
    \midrule
    \multicolumn{4}{l}{\emph{Qwen3-0.6B}} \\
Round-to-nearest & 0.145 & 2.08 & 0.036 \\
Stochastic rounding & 33.235 & 248.47 & 2.346 \\
\oursrow \textbf{Draw-once} & \textbf{0.051} & \textbf{0.93} & 0.031 \\
\addlinespace
\multicolumn{4}{l}{\emph{Qwen3-1.7B}} \\
Round-to-nearest & 0.103 & 3.78 & 0.069 \\
Stochastic rounding & 33.229 & 715.36 & 6.853 \\
\oursrow \textbf{Draw-once} & \textbf{0.041} & \textbf{2.14} & 0.065 \\
 
    \bottomrule
  \end{tabular}
\end{table}

On Qwen3-0.6B, a redraw changes \ChangedSR\% of the weights per export and draw-once changes \ChangedOnce\%. With the same encoding, the mean patch shrinks from \PatchSR{} MB to \PatchOnce{} MB, a factor of \PatchRatio{} (\cref{tab:sync}). In the single-GPU runs, draw-once takes about as long to export as round-to-nearest at both model sizes.
The movement predicted from the master's steps agrees with the observed share to within \TheoryStepDev\% at every export in \TheoryRuns{} draw-once runs on the three-bit power-of-two grid (\cref{app:proofs}).

\begin{figure}[t]
  \centering\includegraphics[width=\columnwidth]{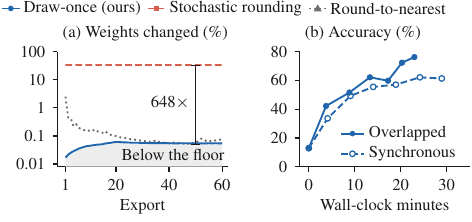}
  \caption{\textbf{With draw-once, sampling overlaps training at \OverlapRatio{} times the synchronous step rate.} Qwen3-0.6B with power-of-two scales. (a) Fraction of weights changed, averaged over three seeds. The shaded region lies below the total-variation floor for unbiased exports. (b) Held-out accuracy against elapsed training time for the synchronous and overlapped draw-once runs at seed 0.}
  \label{fig:sync}
\end{figure}

With one trainer and three samplers, draw-once completes \SyncStepsHr{} steps per hour synchronously. Sampling from the previous export while training with token-level importance weights raises the rate to \OverlapStepsHr{}, a factor of \OverlapRatio{} (\cref{fig:sync}). Mean answer lengths differ by less than one token. Average held-out accuracies in this single-seed pair are \SyncAvg\% and \OverlapAvg\%, respectively (\cref{tab:timing}).
In the timed pair with the clipped surrogate, each sampler receives \WireOnce{} MB on average per draw-once export after the initial load. A redraw ships the full \WireSR{} MB copy. Before generating rollouts, each sampler verifies its synchronized weights against the trainer's checksum for that export (\cref{app:setup}).

\subsection{The Serving Grid and the Released Checkpoint}
\label{sec:rtn}

The starting grid explains the different rankings of round-to-nearest in \cref{tab:main}. On the power-of-two grid many released bfloat16 weights start exactly on ties (\cref{app:static}); a small first update can flip them by a whole cell. On the max-abs grid ties are rare, and round-to-nearest has the highest mean lagged accuracy at both sizes.
The saved checkpoint gives a second test. We score the final served copy and then evaluate the same puzzles after casting its float32 master to bfloat16. In all \OnceReleaseCount{} draw-once runs with paired evaluations, the bfloat16 checkpoint scores within five points of the served copy or higher. These runs cover both FP8 grids and the finer grids tested at 0.6B. All six max-abs round-to-nearest runs exceed that loss threshold (\cref{app:release}).

On a bfloat16 sampler, weights start on grid points. With the same warm-up and learning rate, round-to-nearest reaches \BFRTN\% average accuracy against \BFOnce\% for draw-once. Doubling the learning rate and removing warm-up brings round-to-nearest to \BFTuned\%; quadrupling it produces short answers at all three seeds (\cref{app:late}).

\section{Related Work}
\label{sec:related}

Stochastic rounding makes low-precision arithmetic unbiased~\citep{gupta2015deep,croci2022stochastic}, and \citet{zamirai2020revisiting} show that nearest rounding of the weight update cancels small updates in 16-bit training and that stochastic rounding repairs it.
For RL post-training, \citet{gao2026bf16} identify the withheld updates of the bfloat16 export and propose stochastic rounding or error feedback with the loss computed on the served copy.
\citet{nazemi2026closing} close the convergence gap between bfloat16 and float16 training the same way and report that it helps in asynchronous runs of a 397B-parameter model.
We couple successive exports so that a fixed master always serves the same value. During training the encoded patch stays small at both sizes (\cref{tab:sync}).
Analyses of quantization-aware training on coarse grids take a weight's position in its cell to be uniform~\citep{xu2026beyond}, and our offset creates the phase that a released checkpoint lacks.

A sampler and a trainer that disagree on token probabilities destabilize policy-gradient training.
Recent work removes the numerical part of that gap by running both in float16~\citep{qi2025defeating} or by giving the trainer the sampler's quantized forward pass~\citep{xi2026jetrl,gu2026qarl}, and corrects what remains in the objective~\citep{qiu2026fp8rl,zhou2026ais}.
Asynchronous systems accept stale rollouts for throughput~\citep{espeholt2018impala,li2026staleflow} and control the policy lag with trust regions, importance weights and gradient projection~\citep{yang2026stale,huang2026stable,xu2026gac}.
We quantify the part of the local policy gap due to rounding and minimize it for fixed masters (\cref{prop:lag}).

Shipping only the weights that changed makes frequent synchronization affordable~\citep{miahi2026understanding,hu2026sparserlsync,peng2026arealdte,jiang2026nemodcr}.
These systems measure their sparsity under round-to-nearest.
On a uniform grid, draw-once approaches the lower bound on changes for small master moves averaged over starting phases (\cref{cor:sync}).

\Method{} joins three classical ideas.
A dither that is uniform over one quantizer step makes the mean of the quantization error independent of the input~\citep{schuchman1964dither,wannamaker2000theory}, and our offset is such a dither, drawn once and carried by the signal.
Keeping one random number per unit to hold two random selections together is the method of permanent random numbers in survey sampling~\citep{brewer1972selecting}, where the largest overlap of two samples is a studied problem~\citep{matei2005maximal}, and of rounding under shared randomness in replicable learning~\citep{impagliazzo2022reproducibility,kalavasis2024replicable}.
Its optimality for convex costs is the monotone coupling of optimal transport on the line~\citep{villani2003topics,santambrogio2015optimal}.
In RL post-training, samplers may hold different versions. A single threshold per weight couples every pair of versions and lets us predict how many weights a sampler must receive on average and how much rounding contributes to its local policy divergence (\cref{cor:sync,prop:lag}).

\section{Conclusion}
\label{sec:conclusion}

Two unbiased exports can move a sampler's policy even when the master weights stand still.
\Method{} reuses one random threshold per weight. At every lag, the expected distance between two exports equals the distance between the two fixed master values.
On logic puzzles with a one-export lag, draw-once gained \SmallMaxGain{} to \SmallPowGain{} points of average held-out accuracy over fresh draws. Every matched seed favored it on both grids at both sizes.
On the 0.6B power-of-two grid, draw-once cut encoded patch size by a factor of \PatchRatio{}. With one trainer and three samplers, overlapped draw-once training sustained \OverlapRatio{} times the synchronous step rate.
A random offset at load lets the trainer keep its ordinary cast and send each sampler only the weights that cross a rounding boundary.

\ifdefined\CodeURL
\section*{Code Availability}

The MLX and PyTorch implementations and the numerical checks of the theory are available at \url{\CodeURL}.
\fi

\clearpage
\bibliography{references}
\bibliographystyle{icml2027}

\newpage
\appendix
\onecolumn
\section{Proofs}
\label{app:proofs}

Throughout, fix one weight and a grid that is uniform with cell $s$ near it, and write $x = m/s$.
$\mathrm{SR}(m)$ is the law that puts mass $\frc(x)$ on the upper neighbour $s(\lfloor x \rfloor + 1)$ and the rest on the lower neighbour $s\lfloor x \rfloor$.
Any export that takes values in the two neighbours of $m$ and has mean $m$ has this law.

\subsection{Proof of Proposition~\ref{prop:unbiased}}

For $r \in [0,1)$ we have $\lfloor x + r \rfloor = \lfloor x \rfloor + \mathbf{1}\{r \ge 1 - \frc(x)\}$.
With $r$ uniform and independent of $x$, the indicator has probability $\frc(x)$, so $q = s\lfloor x + r\rfloor$ has the law $\mathrm{SR}(m)$ and mean $m$.

For the converse let $o$ take values in $(-s/2, s/2]$ and write $v = o/s$.
For $x \in [0,1)$, $\rn(m + o)/s = \lfloor x + v + \tfrac12 \rfloor = \mathbf{1}\{v \ge \tfrac12 - x\}$.
Unbiasedness for every such $x$ requires $\Prob(v \ge \tfrac12 - x) = x$ for all $x \in [0,1)$, which is the statement that $v$ is uniform on $(-\tfrac12, \tfrac12]$. \qed

\subsection{Proof of Theorem~\ref{prop:coupling}}

Let $\mu = \mathrm{SR}(m)$ and $\mu' = \mathrm{SR}(m')$, with quantile functions $F_\mu^{-1}$ and $F_{\mu'}^{-1}$.
For $u \in (0,1)$, $F_\mu^{-1}(u)$ is the lower neighbour of $m$ when $u \le 1 - \frc(x)$ and the upper neighbour otherwise.
By the identity in the previous proof, the export~\eqref{eq:threshold} is $q_\star = F_\mu^{-1}(r)$ for almost every $r$, and likewise $q'_\star = F_{\mu'}^{-1}(r)$ with the same $r$.
The pair $(q_\star, q'_\star)$ is therefore the monotone coupling of $\mu$ and $\mu'$.
On the real line the monotone coupling minimizes $\E\, c(Y - X)$ over all couplings $(X, Y)$ of two laws for every non-negative convex $c$, and is the only minimizer when $c$ is strictly convex~\citep[Theorem~2.9]{santambrogio2015optimal}; see also \citet[Section~2.2]{villani2003topics}.
A convex $c$ of either sign differs from a non-negative one by an affine function, whose expectation is the same under every coupling.
A rule whose snapshots have the laws $\mu$ and $\mu'$ produces such a coupling, which gives the inequality.

Both claims of the theorem can also be checked directly and for every convex $c$, because a coupling of two two-point laws has one free parameter.
Take $m \le m'$, let $a$ and $a + s$ be the neighbours of $m$ and $b$ and $b + s$ those of $m'$, and write $\delta = b - a \ge 0$ and $\phi$, $\phi'$ for the two phases.
A coupling is fixed by $p = \Prob(q = a + s,\ q' = b)$, the probability that the first export is rounded up and the second down.
The other pairs (up, up), (down, up) and (down, down) have the probabilities $\phi - p$, $\phi' - \phi + p$ and $1 - \phi' - p$.
On the four pairs the difference $q' - q$ takes the values $\delta - s$, $\delta$, $\delta + s$ and $\delta$, so
\[
  \E\, c(q' - q) \;=\; c(\delta) + (\phi' - \phi)\big[c(\delta + s) - c(\delta)\big] + p\,\big[c(\delta + s) + c(\delta - s) - 2c(\delta)\big].
\]
The last bracket is non-negative for convex $c$ and positive for strictly convex $c$.
The expectation is therefore least at the smallest $p$ that leaves all four probabilities non-negative, $p = \max(0, \phi - \phi')$, and for strictly convex $c$ only there.
Under a kept threshold the first export is rounded up and the second down exactly when $1 - \phi \le r < 1 - \phi'$, which has probability $\max(0, \phi - \phi')$.

For the absolute movement, Jensen's inequality gives $\E|q' - q| \ge |\E q' - \E q| = |m' - m|$ for any such rule.
Under the monotone coupling $q'_\star - q_\star$ has the sign of $m' - m$ for every $r$, so equality holds.
If $|m' - m| = d\,s$ with $d \le 1$, then $q'_\star - q_\star$ is $0$ or $\pm s$, the latter with probability $d$ (\cref{cor:sync}), so $\E(q'_\star - q_\star)^2 = s^2 d = s\,|m' - m|$. \qed

\subsection{Proof of Corollary~\ref{cor:sync}}

Take $x < x'$.
The two exports differ exactly when an integer lies in $(x + r, x' + r]$, which for uniform $r$ has probability $\min(1, x' - x)$.

For the floor, any coupling of $\mu$ and $\mu'$ satisfies $\Prob(q \ne q') \ge \mathrm{TV}(\mu, \mu')$~\citep[Proposition~4.7]{levin2017markov}.
Let the move be $d \le 1$ cells from phase $\phi$.
If $\phi + d \le 1$ the two laws live on the same two points and $\mathrm{TV} = d$.
Otherwise they share only the grid value between them, with masses $\phi$ and $2 - \phi - d$, and $\mathrm{TV} = 1 - \min(\phi,\, 2 - \phi - d)$.
Integrating over a uniform phase,
\[
  \E_\phi \mathrm{TV} \;=\; (1 - d)\,d + \int_{1-d}^{1-d/2} (1 - \phi)\,d\phi + \int_{1-d/2}^{1} (\phi + d - 1)\,d\phi \;=\; d - \frac{d^2}{4}. \qed
\]

\subsection{Proof of Proposition~\ref{prop:lag}}

Unbiased snapshots give $\E\Delta = D$, and independence across weights gives $\E[\Delta_i \Delta_j] = D_i D_j$ for $i \ne j$.
Hence $\E[\Delta^{\!\top} F \Delta] = \sum_{i,j} F_{ij} D_i D_j + \sum_i F_{ii} \Var(\Delta_i)$.
Under \method{} with $d_i \le 1$, $\Delta_i$ is $s_i$ in the direction of the move with probability $d_i$ and zero otherwise, so $\Var(\Delta_i) = s_i^2 d_i (1 - d_i)$.
Under a redraw inside a cell the two exports are independent Bernoulli variables scaled by $s_i$, so $\Var(\Delta_i) = s_i^2[\phi_i(1 - \phi_i) + \phi'_i(1 - \phi'_i)]$, and $\E\,\phi(1-\phi) = \tfrac16$ for a uniform phase.
Since $\Var(\Delta_i) = \E\Delta_i^2 - D_i^2$ and \cref{prop:coupling} with $c(z) = z^2$ minimizes $\E\Delta_i^2$, \method{} minimizes every term of the rounding sum. \qed

\begin{figure}[h]
  \centering
  \includegraphics[width=0.33\textwidth]{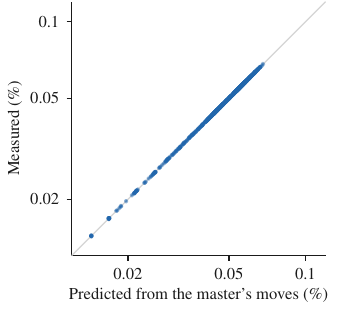}
  \caption{\textbf{\cref{cor:sync} predicts the share of weights that changes at every export.} Each point is one export of one of the \TheoryRuns{} \method{} runs at FP8 precision with power-of-two scales, placed at the share predicted from the master's moves alone and at the share measured; the line is equality. No export departs from the prediction by more than \TheoryStepDev\%.}
  \label{fig:theory}
\end{figure}

\section{How Much of a Step Reaches the Sampler}
\label{app:firstorder}

Let $g$ be the gradient that a step used and $\delta$ the step it took.
The step intends a first-order change of the loss $\sum_i g_i \delta_i$, and the sampler receives $\sum_i g_i (q'_i - q_i)$.
For fixed $g$ and $\delta$ and moves of at most one cell, under \method{} weight $i$ crosses with probability $d_i = |\delta_i|/s_i$ and then moves $s_i$ in the direction of $\delta_i$, independently across weights, so the received change has mean $\sum_i g_i \delta_i$, the whole step, and variance $\sum_i g_i^2 s_i^2 d_i(1 - d_i)$.
For a gradient spread over $N_{\mathrm{eff}} = (\sum_i |g_i|)^2 / \sum_i g_i^2$ weights and moves of $d$ cells the relative standard deviation is about $(d\,N_{\mathrm{eff}})^{-1/2}$.
The rounding error is averaged across the weights of one export, so a copy that changed a twentieth of a percent of its weights still realizes the step.
Under a redraw the variance is $\sum_i g_i^2 s_i^2[\phi_i(1 - \phi_i) + \phi'_i(1 - \phi'_i)]$, which does not shrink with the move.

Evaluated at seed 0 on the gradients and moves of Qwen3-0.6B, the relative standard deviation of the delivered step is \FOOnceBF\% on bfloat16 and \FOOnceThree\% at FP8 precision under \method{}, against \FOSRBF\% and \FOSRThree\% for a redraw of the same masters.
At FP8 precision the predicted standard deviation for a redraw exceeds the intended first-order change.

\section{The Clipped Surrogate Under a Noisy Ratio}
\label{app:clip}

\cref{fig:mech}(b) measures the token ratios that enter the clipped objective, and this section derives what the clipped surrogate does with the rounding term.

PPO and GRPO give a sampled token the term $\min\!\big(\rho A,\ \mathrm{clip}(\rho, 1 - \epsilon, 1 + \epsilon)\,A\big)$, where $\rho$ is the ratio of the trained policy's probability to the sampler's and $A$ is the advantage of the token's answer.
A token whose ratio is outside the band has a gradient only when its advantage points against the direction in which the ratio left.
Split the gradient $G(A)$ of the loss over a batch into the part that is odd in the advantages and the part that is even:
\[
  \text{signal} = \tfrac12\big(G(A) - G(-A)\big), \qquad \text{pull} = \tfrac12\big(G(A) + G(-A)\big).
\]
For a plain policy gradient and for importance-weighted objectives the gradient is odd and the pull is zero.
For the clipped surrogate
\[
  \text{pull} \;=\; \frac{1}{N} \sum_{\text{tokens outside the band}} \frac{|A|}{2}\,\rho\,\sigma\,\nabla \log \pi(a),
\]
with $N$ the number of answer tokens and $\sigma = +1$ for a token the trained policy rates above the band and $-1$ below.
The pull contains the size of each advantage and not its sign: its direction for each token is independent of the sign of its advantage.
A descent step along it lowers the tokens that the trained copy rates higher than the sampler did and raises the ones it rates lower, a step toward the sampler's policy.

When the ratio is the learned drift between two policies, this is the trust region of the objective.
When the ratio is the rounding noise between two redrawn exports, the policy being pulled toward is one random draw, replaced by another at the next step.
A redraw can move tokens outside the clip band even if the learning rate is zero.
In the lagged Qwen3-0.6B runs with power-of-two scales, the mean fraction of response tokens whose gradient the clip zeroed was \LagPowSRClip\% under a redraw and \LagPowOnceClip\% under draw-once (\cref{sec:stale}).
With a redraw at seed 0, narrowing the clip band to $\epsilon=0.1$ gives \ClipNarrow\% average accuracy and widening it to $0.4$ gives \ClipWide\%. Halving the learning rate at the original width gives \ClipHalfRate\%. All three remain below draw-once at the original settings (\cref{tab:depth}).

\section{Drawing the Threshold After a Warm-Up}
\label{app:late}

The offset of \cref{eq:offset} can be applied to any master, including one that has left the grid.
Let $m$ be a master weight at export $K$, anywhere in its cell, and $u$ uniform on $[-s/2, s/2)$ and independent of it.
Then $\rn(m + u)$ has the law $\mathrm{SR}(m)$, so export $K$ is one more unbiased draw, and $\frc((m+u)/s)$ is uniform whatever $m$ was, so from export $K$ on \cref{prop:unbiased,prop:coupling,prop:lag} apply as they do from load.

This gives a stack that already ships stochastic rounding a migration path.
Seeding its draw by $\min(t, K)$ instead of by $t$ freezes the thresholds after export $K$ and yields \cref{eq:threshold}.
Adding the frozen draw to the master once at export $K$ and switching the export to round-to-nearest yields \cref{eq:offset}, with a master that can be cast to any format.

The offset can also be drawn after warm-up. \Cref{tab:bf16} compares this choice with an offset at load and with two learning rates for round-to-nearest. Drawing at load has the higher mean accuracy of the two draw-once variants.
\begin{table}[h]
  \caption{\textbf{Warm-up and learning rate change the bfloat16 baseline.} Qwen3-0.6B, synchronous knights-and-knaves, three seeds. Weight changes average all 60 exports. Patch sizes average the logged diagnostics.}
  \label{tab:bf16}
  \centering
  \small
  \begin{tabular*}{\linewidth}{@{\extracolsep{\fill}}lrrrr@{}}
    \toprule
    Export rule & Average acc. (\%) & Final acc. (\%) & Weights changed (\%) & Patch (MB) \\
    \midrule
    Round-to-nearest & 39.9 & 57.3 & 0.51 & 8.6 \\
Stochastic rounding & 55.1 & 72.5 & 7.67 & 71.8 \\
\textbf{Draw-once} & 57.4 & 69.9 & 0.62 & 10.5 \\
Draw-once after warm-up & 54.8 & 70.3 & 1.56 & 22.4 \\
Round-to-nearest, $2\times$ rate, no warm-up & 61.1 & 74.6 & 1.20 & 18.5 \\
Round-to-nearest, $4\times$ rate, no warm-up & 28.0 & 29.7 & 1.76 & 25.1 \\
 
    \bottomrule
  \end{tabular*}
\end{table}

\section{Experimental Detail}
\label{app:setup}

\paragraph{Models.}
We train all parameters of Qwen3-0.6B, Qwen3-1.7B and SmolLM2-360M from their released checkpoints. The main comparison uses the two Qwen3 models, and the arithmetic experiment also tests SmolLM2.

\paragraph{Formats.}
A bfloat16 sampler holds the bfloat16 cast of every tensor.
An FP8-precision sampler holds every matrix rounded to 3 mantissa bits and every other tensor in bfloat16.
On the grid with power-of-two scales each weight keeps the top 3 mantissa bits of its own binade.
MXFP8 stores E4M3 elements under one power-of-two scale per block of 32 weights.
Its cell equals this grid's cell wherever the element is a normal number below the largest finite value, which holds for \SameCellLow\% to \SameCellHigh\% of the weights of the checkpoints in \cref{tab:census}.
For max-abs scaling, the tensor maximum at load sets the scale. We take the significand $c\in[1,2)$ of that maximum divided by 448, round $w/c$ to three mantissa bits and multiply by $c$. The training code emulates these grids in floating point.
Activations are not quantized in either case, so the comparison isolates the weight export.

\paragraph{Sampling and lag.}
Answers have at most 512 new tokens and are sampled at temperature 1.
With a lag of $L$ the answers used at step $t$ are sampled from the export of step $\max(0,t-L)$, and their log-probabilities under that copy are kept for the objective.
The trainer's forward and backward pass run on the current served copy.

\paragraph{Objectives.}
Advantages are the reward of an answer minus the mean of its group, divided by the group's standard deviation.
The clipped surrogate uses $\epsilon = 0.2$.
Token-level truncated importance weights cap each token's ratio at 2.

\paragraph{Implementations and timing.}
The PyTorch implementation extends the served-copy training loop of \citet{gao2026bf16}. Training and checkpoint probes use NVIDIA A100 80 GB PCIe GPUs. The independent MLX implementation runs on an Apple M5 Max.
In \cref{tab:sync}, each training run uses one GPU. Its export timer covers rounding the weights and counting the entries that changed. We omit the first five exports, average the remaining 55, then average the three seed means. These times describe the training runs, which could share a GPU with another run.

\Cref{tab:timing} uses a dedicated four-GPU node with one trainer and three samplers. Here the export timer also includes handing the new weights to the sampler queue. Each sampler verifies its copy against the trainer's. We measure throughput between consecutive exports after the first four intervals, including a local checkpoint write every tenth step for deferred evaluation. Omitting those intervals gives a speed ratio of \OverlapRatioPlain{} for overlapped against synchronous training. We use one run per configuration under the same storage conditions.

\begin{table}[ht]
  \caption{\textbf{Overlapped draw-once training processes more steps per hour than synchronous draw-once training.} Qwen3-0.6B, one trainer and three samplers, seed 0. Tokens per answer average all training steps. Output length varies with the learned policy and affects the step rate. Transfer size is per sampler after the initial load. Accuracy averages the six post-start evaluations.}
  \label{tab:timing}
  \centering\small
  \begin{tabular*}{\linewidth}{@{\extracolsep{\fill}}llrrrrr@{}}\toprule
    Export rule & Sampling and objective & Steps/h & Tokens/answer & MB/sync & Export (s) & Accuracy (\%) \\\midrule
    Draw-once & Synchronous & 122.7 & 262 & 2.86 & 0.234 & 53.1 \\
Draw-once & Lag 1, importance weights & 154.9 & 263 & 2.86 & 0.234 & 60.7 \\
Draw-once & Lag 1, clipped & 158.5 & 258 & 2.74 & 0.236 & 53.4 \\
Stochastic & Lag 1, clipped & 226.6 & 111 & 1,192.20 & 3.269 & 35.6 \\
Round-to-nearest & Lag 1, clipped & 390.4 & 70 & 11.81 & 0.253 & 32.8 \\
 
    \bottomrule
  \end{tabular*}
\end{table}

\paragraph{Patches.}
Encoded sizes measure the same position-and-value encoding for every rule. In the main comparison the redraw logs patch sizes at exports 1, 2, 5, 10 and every tenth export thereafter. Draw-once and round-to-nearest log every export. Actual transfers use the implementation's sparse patch for draw-once and a dense copy for a redraw. We report mean bytes per receiving sampler after the first dense load. Patch sizes use decimal megabytes.

\paragraph{Seeds.}
A seed sets the sampling of rollouts and the randomness of the export.
The sequence of training problems is fixed.

\raggedbottom
\section{Measurements on Released Checkpoints}
\label{app:static}

For each checkpoint, three stochastic exports each sample answers to the same 32 prompts. \Cref{tab:static,tab:noise} average token divergence and the share of ratios outside $[0.8,1.2]$ over the six ordered pairs of exports. The tie counts in \cref{tab:static,tab:census} come directly from the nonzero entries of the checkpoint matrices.

\begin{table}[H]
  \caption{\textbf{Two costs of current exports are visible in released checkpoints before any training.} The redraw columns give the policy distance between two stochastic-rounding exports of one \emph{unchanged} checkpoint and the share of sampled tokens whose probability ratio leaves $[0.8, 1.2]$ (linear layers at FP8 precision, 3 exports per model, mean over pairs). The last column gives the share of matrix weights that sit exactly on a round-to-nearest tie of MXFP8.}
  \label{tab:static}
  \centering
  \small
  \begin{tabular*}{\linewidth}{@{\extracolsep{\fill}}lrrrr@{}}
    \toprule
    & & \multicolumn{2}{c}{Redraw, same weights} & \\
    \cmidrule(lr){3-4}
    Model & Size (B) & Nats/token & Outside (\%) & Ties (\%) \\
    \midrule
    Qwen3-0.6B & 0.6 & 0.0285 & 16.3 & 6.19 \\
Qwen3-1.7B & 1.7 & 0.0230 & 9.8 & 6.20 \\
Qwen3-4B & 4.0 & 0.0116 & 7.4 & 6.19 \\
Qwen3-8B & 8.2 & 0.0081 & 8.5 & 6.19 \\
Qwen3-14B & 14.8 & 0.0083 & 8.5 & 6.19 \\
 
    \bottomrule
  \end{tabular*}
\end{table}

On Qwen3-0.6B the divergence between two redrawn exports is \NoiseBitsFive, \NoiseBitsFour{} and \NoiseBitsThree{} nats per token at 5, 4 and 3 mantissa bits on all matrices, consistent with the quadratic dependence on grid spacing in \cref{prop:lag}.

\begin{table}[H]
  \caption{\textbf{A redraw moves the policy of every model tested.} Token divergence and ratios outside $[0.8,1.2]$ under two FP8 layouts. The released layout quantizes linear layers in $128\times128$ blocks and keeps the embedding and output head in bfloat16. The other uses power-of-two scales on all matrices.}
  \label{tab:noise}
  \centering
  \small
  \begin{tabular*}{\linewidth}{@{\extracolsep{\fill}}lr rr rr@{}}
    \toprule
    & & \multicolumn{2}{c}{Released layout} & \multicolumn{2}{c}{All matrices} \\
    \cmidrule(lr){3-4}\cmidrule(l){5-6}
    Model & Params (B) & Nats/token & Outside (\%) & Nats/token & Outside (\%) \\
    \midrule
    Qwen3-0.6B & 0.60 & 0.0285 & 16.3 & 0.0306 & 17.9 \\
Qwen3-1.7B & 1.72 & 0.0230 & 9.8 & 0.0250 & 10.9 \\
Qwen3-4B & 4.02 & 0.0116 & 7.4 & 0.0142 & 9.1 \\
Qwen3-8B & 8.19 & 0.0081 & 8.5 & 0.0097 & 8.7 \\
Qwen3-14B & 14.77 & 0.0083 & 8.5 & 0.0098 & 8.6 \\
Qwen2.5-0.5B-Instruct & 0.49 & 0.0372 & 30.1 & 0.0558 & 37.1 \\
Qwen2.5-7B-Instruct & 7.62 & 0.0165 & 10.9 & 0.0132 & 10.2 \\
SmolLM2-360M-Instruct & 0.36 & 0.0278 & 21.5 & 0.0526 & 30.9 \\
SmolLM2-1.7B-Instruct & 1.71 & 0.0281 & 20.5 & 0.0615 & 29.2 \\
 
    \bottomrule
  \end{tabular*}
\end{table}

\begin{table}[H]
  \caption{\textbf{About one in sixteen nonzero matrix weights lies on an MXFP8 tie in each of the \CensusModels{} checkpoints.} Ties are exact midpoints between adjacent grid values. The same-cell column gives the share of weights whose MXFP8 cell matches our power-of-two grid.}
  \label{tab:census}
  \centering
  \small
  \begin{tabular*}{\linewidth}{@{\extracolsep{\fill}}lrrrrr@{}}
    \toprule
    & & \multicolumn{2}{c}{MXFP8} & \multicolumn{2}{c}{Ties on other grids (\%)} \\
    \cmidrule(lr){3-4}\cmidrule(l){5-6}
    Checkpoint & Weights (B) & Ties (\%) & Same cell (\%) & MXFP4 & Max-abs \\
    \midrule
    Qwen2.5-0.5B-Instruct & 0.49 & 6.20 & 99.28 & 1.15 & 0.42 \\
Qwen2.5-7B-Instruct & 7.62 & 6.20 & 99.28 & 1.26 & 0.42 \\
Qwen3-0.6B & 0.75 & 6.19 & 99.12 & 1.15 & 0.22 \\
Qwen3-1.7B & 2.03 & 6.20 & 99.31 & 1.16 & 0.22 \\
Qwen3-4B & 4.02 & 6.19 & 99.14 & 1.17 & 0.48 \\
Qwen3-8B & 8.19 & 6.19 & 99.05 & 1.14 & 0.41 \\
Qwen3-14B & 14.77 & 6.19 & 99.09 & 1.14 & 0.54 \\
SmolLM2-360M-Instruct & 0.36 & 6.20 & 99.20 & 1.16 & 0.34 \\
SmolLM2-1.7B-Instruct & 1.71 & 6.19 & 99.06 & 1.16 & 0.70 \\
 
    \bottomrule
  \end{tabular*}
\end{table}

\section{Further Results}
\label{app:more}

\subsection{Seeds and Precision}

On both FP8 grids at both sizes, the lowest draw-once accuracy exceeds the highest redraw accuracy (\cref{tab:seeds}).

\begin{table}[H]
  \caption{\textbf{Draw-once rounding exceeds fresh draws at each matched seed.} Average held-out accuracy on knights-and-knaves with FP8 weights and a one-export lag. Boldface uses the same tolerance as \cref{tab:main}.}
  \label{tab:seeds}
  \centering\small
  \begin{tabular*}{\linewidth}{@{\extracolsep{\fill}}llrrrrrrrr@{}}\toprule
    & & \multicolumn{4}{c}{Qwen3-0.6B} & \multicolumn{4}{c}{Qwen3-1.7B} \\
    \cmidrule(lr){3-6}\cmidrule(l){7-10}
    FP8 scales & Export rule & Seed 0 & Seed 1 & Seed 2 & Mean & Seed 0 & Seed 1 & Seed 2 & Mean \\\midrule
    Power-of-two & Round-to-nearest & 0.0 & 5.3 & 19.1 & 8.1 & 22.3 & 0.0 & 1.1 & 7.8 \\
 & Stochastic rounding & 31.6 & 34.9 & 36.4 & 34.3 & 59.7 & 60.2 & 58.1 & 59.3 \\
 & \textbf{Draw-once} & 58.9 & 54.8 & 47.5 & \textbf{53.7} & 71.8 & 71.1 & 71.9 & \textbf{71.6} \\
\addlinespace
Max-abs & Round-to-nearest & 58.1 & 47.5 & 56.6 & \textbf{54.1} & 72.7 & 73.2 & 69.1 & \textbf{71.7} \\
 & Stochastic rounding & 46.7 & 36.3 & 37.6 & 40.2 & 58.4 & 57.8 & 60.5 & 58.9 \\
 & \textbf{Draw-once} & 57.6 & 49.6 & 46.8 & \textbf{51.3} & 70.3 & 71.8 & 70.3 & 70.8 \\
 
    \bottomrule
  \end{tabular*}
\end{table}

\begin{table}[H]
  \caption{\textbf{The policy gap from a redraw shrinks on finer weight grids.} Qwen3-0.6B, one export behind, seed 0, clipped surrogate. Accuracy is the mean over training with the final score in parentheses. Divergence is measured on sampled tokens.}
  \label{tab:precision}
  \centering\small
  \begin{tabular*}{\linewidth}{@{\extracolsep{\fill}}lrr rr@{}}\toprule
   & \multicolumn{2}{c}{Accuracy (\%)} & \multicolumn{2}{c}{Nats per token}\\
   \cmidrule(lr){2-3}\cmidrule(l){4-5}
   Mantissa bits & Stochastic rounding & \textbf{Draw-once} & Stochastic rounding & \textbf{Draw-once}\\\midrule
   3 (FP8) & 31.6 (43.4) & 58.9 (73.4) & 0.0271 & 0.0001 \\
4 & 49.0 (61.3) & 58.8 (69.5) & 0.0074 & 0.0001 \\
5 & 49.9 (58.6) & 52.3 (65.2) & 0.0020 & 0.0001 \\
7 (bfloat16) & 59.0 (73.8) & 57.1 (77.7) & 0.0002 & 0.0001 \\
 
   \bottomrule
  \end{tabular*}
\end{table}

At seed 0 the accuracy gap narrows on finer grids and reverses on bfloat16 (\cref{tab:precision}).

\subsection{Deeper Lags and Other Objectives}
\label{app:depth}

\begin{table}[H]
  \caption{\textbf{One export behind, draw-once gains most over fresh draws with the clipped surrogate.} Qwen3-0.6B on knights-and-knaves with power-of-two FP8 scales, seed 0. Entries give average held-out accuracy over training in percent, with final accuracy in parentheses. Importance weights are capped at 2.}
  \label{tab:depth}
  \centering\small
  \begin{tabular*}{\linewidth}{@{\extracolsep{\fill}}clrr@{}}\toprule
   Lag (exports) & Objective & Stochastic rounding & \textbf{Draw-once}\\\midrule
   1 & Clipped surrogate & 31.6 (43.4) & 58.9 (73.4) \\
1 & Plain policy gradient & 54.9 (69.5) & 58.5 (69.9) \\
1 & Importance weights & 55.0 (69.1) & 58.5 (72.7) \\
4 & Importance weights & 47.6 (60.5) & 51.9 (60.2) \\
 
   \addlinespace[0.7em]
   Lag (exports) & Export rule & Clipped surrogate & Importance weights\\\midrule
   4 & \textbf{Draw-once} & 46.4 (58.2) & 51.9 (60.2) \\
16 & \textbf{Draw-once} & 34.2 (32.8) & 40.1 (53.5) \\
 
   \bottomrule
  \end{tabular*}
\end{table}

With draw-once four or sixteen exports behind, token-level importance weights yield higher average and final accuracy than clipping at the same lag. \Cref{app:clip} reports the clip-width and learning-rate controls.

\subsection{A Second Task and a Second Model Family}

Arithmetic uses 32 prompts per step with eight answers of at most 64 tokens, 200 optimizer steps and no warm-up. Comparisons in \cref{tab:other} are within each model's task configuration.

\begin{table}[H]
  \caption{\textbf{A kept threshold also improves lagged accuracy on integer arithmetic.} Mean accuracy across eight evaluations and final accuracy in parentheses. Each cell is one run at seed 0, with power-of-two scales and the clipped surrogate.}
  \label{tab:other}
  \centering\small
  \begin{tabular*}{\linewidth}{@{\extracolsep{\fill}}l rr rr@{}}\toprule
   & \multicolumn{2}{c}{No lag} & \multicolumn{2}{c}{One export behind}\\
   \cmidrule(lr){2-3}\cmidrule(l){4-5}
   Model & Stochastic rounding & \textbf{Draw-once} & Stochastic rounding & \textbf{Draw-once}\\\midrule
   Qwen3-0.6B & 83.3 (84.0) & 81.7 (82.0) & 77.3 (73.4) & 81.8 (84.4) \\
SmolLM2-360M & 63.3 (62.1) & 65.0 (69.5) & 62.2 (62.1) & 65.0 (68.8) \\
 
   \bottomrule
  \end{tabular*}
\end{table}

\clearpage
\subsection{The Released Checkpoint}
\label{app:release}

\begin{figure}[h]
  \centering
  \includegraphics{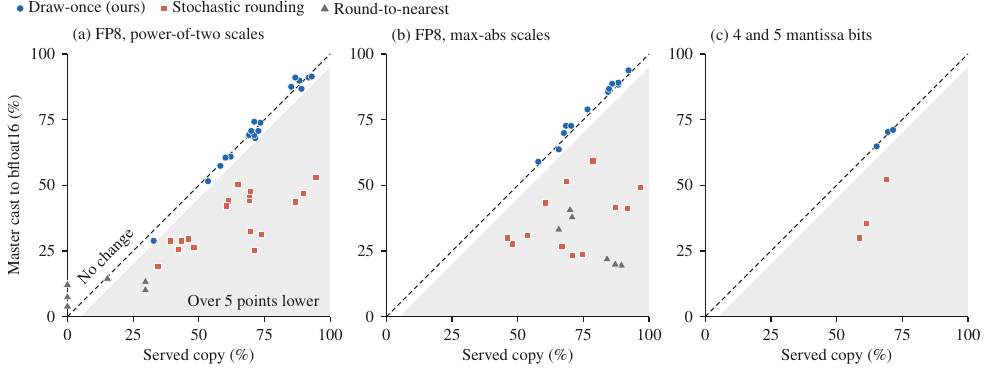}
  \caption{\textbf{All \OnceReleaseCount{} draw-once runs stay above the five-point loss threshold when the master is cast to bfloat16.} Each point is one single-card 60-step run of either Qwen3 model, the runs of \cref{tab:release}: final accuracy of the served copy against that of the float32 master cast to bfloat16 by round-to-nearest, on (a) FP8 with power-of-two scales, (b) FP8 with max-abs scales and (c) grids of 4 and 5 mantissa bits. On the dashed diagonal the cast changes nothing, and the shaded region is more than five points below it.}
  \label{fig:release}
\end{figure}

For each run we evaluate the final served copy, the float32 master and its bfloat16 cast on the same held-out questions. \Cref{fig:release} shows the paired scores, for each grid. \Cref{tab:release} gives the means and counts separately by model and grid.

\begin{table}[h]
  \caption{\textbf{No draw-once run loses more than five points when the master is cast to bfloat16.} Final accuracy in percent of the served copy, of its float32 master, and of that master cast to bfloat16 by round-to-nearest, over the single-card 60-step runs on grids coarser than bfloat16 (all lags and objectives). The last column counts runs whose cast is more than five points below the served copy.}
  \label{tab:release}
  \centering
  \small
  \begin{tabular*}{\linewidth}{@{\extracolsep{\fill}}llcrrrl@{}}
    \toprule
    Sampler format & Export rule & Runs & Served copy & float32 master & bfloat16 cast & Cast lower \\
    \midrule
    \multicolumn{7}{l}{\emph{Qwen3-0.6B}} \\
Power-of-two & Round-to-nearest & 3 & 15.0 & 23.7 & 13.3 & 1 of 3 \\
Power-of-two & Stochastic rounding & 12 & 54.0 & 54.2 & 36.2 & 12 of 12 \\
\oursrow Power-of-two & \textbf{Draw-once} & 12 & 63.8 & 63.2 & 62.9 & 0 of 12 \\
Max-abs & Round-to-nearest & 3 & 68.8 & 68.9 & 37.2 & 3 of 3 \\
Max-abs & Stochastic rounding & 6 & 59.2 & 59.6 & 40.5 & 6 of 6 \\
\oursrow Max-abs & \textbf{Draw-once} & 6 & 67.7 & 67.8 & 69.5 & 0 of 6 \\
4 mantissa bits & Stochastic rounding & 2 & 65.0 & 63.9 & 43.9 & 2 of 2 \\
\oursrow 4 mantissa bits & \textbf{Draw-once} & 2 & 70.5 & 68.8 & 70.7 & 0 of 2 \\
5 mantissa bits & Stochastic rounding & 1 & 58.6 & 61.7 & 30.1 & 1 of 1 \\
\oursrow 5 mantissa bits & \textbf{Draw-once} & 1 & 65.2 & 64.1 & 64.8 & 0 of 1 \\
\addlinespace
\multicolumn{7}{l}{\emph{Qwen3-1.7B}} \\
Power-of-two & Round-to-nearest & 3 & 9.9 & 30.3 & 7.2 & 1 of 3 \\
Power-of-two & Stochastic rounding & 6 & 80.9 & 81.6 & 38.8 & 6 of 6 \\
\oursrow Power-of-two & \textbf{Draw-once} & 6 & 89.0 & 88.5 & 89.6 & 0 of 6 \\
Max-abs & Round-to-nearest & 3 & 86.8 & 83.2 & 20.4 & 3 of 3 \\
Max-abs & Stochastic rounding & 6 & 81.2 & 83.4 & 34.4 & 6 of 6 \\
\oursrow Max-abs & \textbf{Draw-once} & 6 & 87.3 & 87.5 & 88.7 & 0 of 6 \\
 
    \bottomrule
  \end{tabular*}
\end{table}

\end{document}